\documentclass[letterpaper, 10 pt, conference]{ieeeconf}  

\IEEEoverridecommandlockouts                              

\usepackage{graphics} 
\usepackage{graphicx}
\usepackage{epsfig} 
\usepackage{amsmath} 
\usepackage{amssymb}  
 \usepackage{algorithm}
\usepackage{algpseudocode}
\algnewcommand\AAND{\textbf{ and }}
\algnewcommand\Or{\textbf{ or }}
\usepackage{color}
\usepackage{citesort}
\usepackage{url}

\DeclareMathAlphabet{\pazocal}{OMS}{zplm}{m}{n}

\usepackage{array}
\newcolumntype{C}[1]{>{\centering\arraybackslash}p{#1}}
\newcolumntype{M}[1]{>{\raggedright\arraybackslash}p{#1}}

\usepackage{array} 
\newcolumntype{L}[1]{>{\raggedright\let\newline\\\arraybackslash\hspace{0pt}}m{#1}}	
\newcolumntype{S}[1]{>{\centering\let\newline\\\arraybackslash\hspace{0pt}}m{#1}}
\newcolumntype{R}[1]{>{\raggedleft\let\newline\\\arraybackslash\hspace{0pt}}m{#1}}

\title{\LARGE \bf
Towards Robotically Supported Decommissioning of Nuclear Sites
}

\author{Frank Mascarich$^1$, Taylor Wilson$^2$, Tung Dang$^1$, Shehryar Khattak$^1$, Christos Papachristos$^1$, and Kostas Alexis$^1$
\thanks{This material is based upon work supported by the Department of Energy under Award Number [DE-EM0004478].}
\thanks{$^1$ The authors are with the Autonomous Robots Lab, University of Nevada, Reno, 1664 N. Virginia, 89557, Reno, NV, USA
        {\tt\small kalexis@unr.edu} }%
\thanks{$^2$ The author is with the Wilson Laboratory for Radiaton Physics, University of Nevada, Reno, 1664 N. Virginia, 89557, Reno, NV, USA}
\thanks{This paper is presented at the Autonomous Structural Monitoring and Maintenance using Aerial Robots Workshop, IEEE International Conference on Robotics and Automation (ICRA) 2017, Singapore, May 29, 2017}
}

\begin{document}

\maketitle
\thispagestyle{empty}
\pagestyle{empty}

\begin{abstract}
This paper overviews certain radiation detection, perception, and planning challenges for nuclearized robotics that aim to support the waste management and decommissioning mission. To enable the autonomous monitoring, inspection and multi--modal characterization of nuclear sites, we discuss important problems relevant to the tasks of navigation in degraded visual environments, localizability--aware exploration and mapping without any prior knowledge of the environment, as well as robotic radiation detection. Future contributions will focus on each of the relevant problems, will aim to deliver a comprehensive multi--modal mapping result, and will emphasize on extensive field evaluation and system verification.
\end{abstract}


\section{INTRODUCTION}\label{sec:intro}

A history of nuclear research, power generation and military developments has left a legacy of nuclear sites, now requiring careful decommissioning. In the U.S., the goal is that of safe cleanup of the Manhattan Project nuclear sites, the ensuing Cold War nuclear arms race, and the early years of federal nuclear science research and technology development. Sub--tasks of this broad mission include a) nuclear facility decommissioning, b) soil and water cleanup, c) liquid radioactive waste processing and disposition, d) solid radioactive waste treatment, storage and disposal, as well as e) nuclear materials and spent nuclear fuel management. In these challenging tasks requiring careful inspection, characterization, decommissioning and maintenance, robotic systems can be of unparalleled value. Specialized, robustly autonomous, robotic systems that can deal with the dirty, dull, dangerous and difficult environments of the nuclear sites are now required. 

\begin{figure}[h!]
\centering
  \includegraphics[width=0.99\columnwidth]{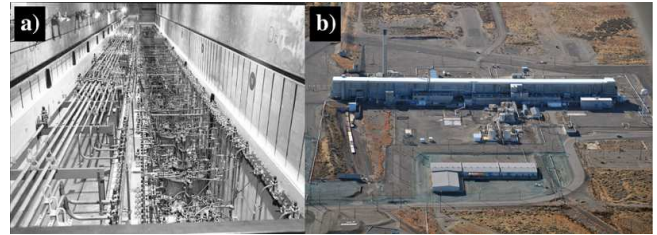}
\caption{Indicative facilities of interest: a) H--Canyon and b) PUREX.}
\label{fig:facilities}
\end{figure}

However, to facilitate the vision of broad and reliable robotic support of the waste management and decommissioning efforts, a set of challenges have to be addressed. Among others this includes the need for pioneering platform designs presenting ultimate mobility, robust autonomy in often visually--degraded and GPS--denied environments, high--resolution mapping and semantic classification, radiation detection and its fusion with multi--modal maps, as well as radiation source localization. Despite very important efforts of the community, such as those described in~\cite{qian2012small,han2013low,bogue2011robots,cortez2009distributed,dudar1999mobile,westrom1994radiation,michal2009red}, a variety of complex problems are yet to be addressed so that robots can operate autonomously in the sites relevant to the decommissioning effort and provide comprehensive mapping and characterization. Indeed, the complexity and the degraded conditions in the facilities of interest are unique to the domain. Figure~\ref{fig:facilities} presents photos of relevant sites.

%
%

In this paper we discuss the problem of robotically supported nuclear waste management and decommissioning in the sense of autonomous exploration, inspection and characterization of the nuclear sites. In particular, our goal is to discuss some of the sensing, path planning, control, system design and implementation challenges that are particular to the environments and mission goals of nuclear site characterization.

\section{CHALLENGES FOR NUCLEARIZED ROBOTICS}\label{sec:challenges}

Nuclear facilities present a unique set of challenges that make robotic support attractive. The most unique and obvious challenge encountered at nuclear facilities is ionizing radiation, ranging from a few times the natural background, to dose rates exceeding many Sv/hr at sites housing spent nuclear fuel, reprocessed material, and also at nuclear accident sites~\cite{groves1983now,kelly2009manhattan,gephart2003hanford,tsubokura2012internal}. This radiation creates an environment hazardous to worker entry, and is in many cases not well characterized because of the lack of human surveys. Loose or airborne contamination creates an internal exposure hazard and penetrating beta, photon and neutron fields pose external exposure threats. Robotic platforms can accurately map radiation fields in environments where dose--rates make personnel entry impossible, and additionally can map radiation and radionuclide contamination in less hazardous environments more efficiently than traditional surveys, keeping with the health physics principle of ALARA (As Low As Reasonably Achievable). Furthermore, nuclear facilities and especially those relevant to the decommissioning mission often are only documented in historic reports. Although rich documentation is available and useful information can be extracted from it, for most of these sites prior maps are not available and capability for GPS--denied operation is required. 

In very high radiation areas, robotics have a unique role in performing tasks that are inaccessible or particularly challenging to humans. Indeed there are operational dose limits on robotics as well. However these limits are much higher and can be mitigated with a combination of radiation hardened electronics, additional shielding, and path planning to minimize time of exposure and dose rates. Radiation damage to semiconductor electronics represents the soft--point for nuclearized robotics, and for very high radiation dose rates, hardened electronics should be chosen, with a wealth of knowledge available from the use and evaluation of electronics packages for space applications. Additionally, work should be conducted for the effect of ionizing radiation on a range of optical systems, including cameras, LiDAR and other proximity/ranging systems. For certain modalities, such as CCD and CMOS optical sensors the effects of ionizing radiation have been demonstrated~\cite{hopkinson2004radiation,goiffon2008ionizing}, yet a comprehensive study on the effects of radiation on many common other robotically--employed sensors is yet to be conducted in order to evaluate the performance of each and identify points of failure. This should be further coupled with the specific task the sensors are used for (e.g. SLAM). 

Robotics for facilities with unsealed sources of radiation should make consideration for ease of decontamination, or in some cases design systems or components to be replaceable or disposable. Additionally, some robotics for nuclear applications will enter areas of high radiation that will also contain a strong thermal source, so temperature ratings and cooling scenarios should be addressed for these applications. This challenges aspects of the design and especially the battery system. Furthermore, mapping and localization of lost or orphan sources~\cite{jarman2011bayesian,towler2012radiation,morelande2009radiological,christie2016radiation} represents a real, as demonstrated by recent source recovery operations. Along with lowering personnel doses in such operations, an autonomous system can optimize a more efficient search method for multi--source localization. 
 
Finally, it is noted that the development of nuclearized robotics for inspection operations, decommissioning, and accident response could have further applications outside the scope of this paper. Autonomous robotics with radiation detection and mapping systems could represent an effective safeguard tool for nuclear facilities and international inspectors to accurately inventory nuclear materials and safeguard against diversions. Such autonomous robotics could provide round--the--clock inspection and inventory of facilities with large layouts and quantities of material. Additionally, lessons learned in the field of nuclearized robotics could provide valuable design feedback for developing spacecraft systems for missions outside of Low Earth Orbit (LEO), where such systems will encounter high radiation fields, and derive usable mission data from radiation sensing.

\section{AUTONOMOUS OPERATION IN DEGRADED VISUAL ENVIRONMENTS}\label{sec:dve}

Nuclearized robotics will be requested to operate in all sorts of challenging environments. Going beyond the current state--of--the--art in robotics for the nuclear domain, autonomous operation (as opposed to teleoperation) and mission--execution in GPS--denied environments will become common. Even more challenging, it is noted that many important applications (e.g. decommissioning) often take place in Degraded Visual Environments (DVEs). Iconic examples include the inspection of the PUREX tunnels and H--Canyon. 

\begin{figure}[h!]
\centering
  \includegraphics[width=0.99\columnwidth]{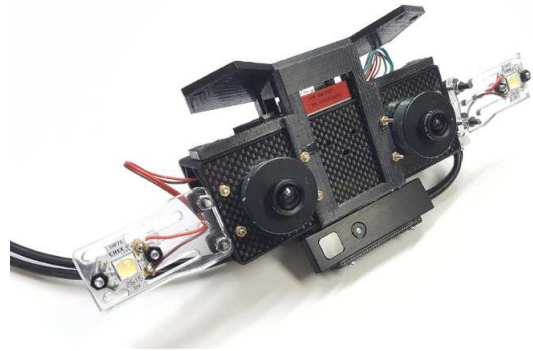}
\caption{Photo of a prototype of the Multi--Modal Mapping Unit.}
\label{fig:m3uphoto}
\end{figure}

\begin{figure}[h!]
\centering
  \includegraphics[width=0.99\columnwidth]{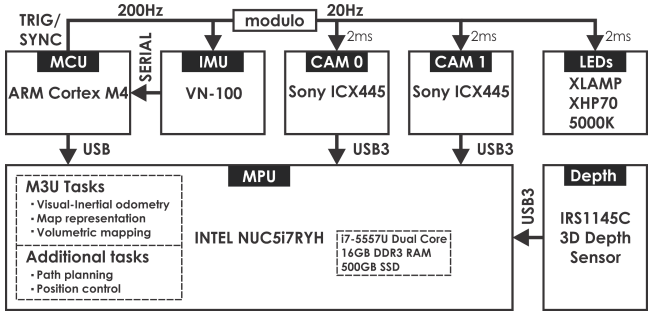}
\caption{Overview of the design diagram of the Multi--Modal Mapping Unit architecture. The microcontroller unit (MCU) is responsible for the visual--inertial subsystem triggering, while a powerful high-level main processing unit (MPU) handles all the data acquisition and processing tasks.}
\label{fig:m3udesign}
\end{figure}

For the problem of autonomous navigation in DVEs, a robust localization and intelligent path planning strategy has to be facilitated. Recent work of our team aims to address the problem through a) multi--modal sensor fusion~\cite{NIRdepth_ICUAS_2017}, and b) localization uncertainty--aware Receding Horizon Exploration and Mapping (RHEM) path planning~\cite{RHEM_ICRA_2017}. In this approach, data from visual cameras synchronized with flashing LEDs (or Near Infrared cameras) are fused with inertial sensor cues and a depth sensor in order to enable robust operation in darkness. As given certain environments, a sensing modality can become ill--conditioned, a multi--modal sensor fusion approach can robustify the overall robot operation and also provide mapping results of higher resolution and fidelity. Towards robotic operation in the complex environments relevant to the nuclear decommissioning effort, our team develops a Multi--Modal Mapping Unit (M3U) a prototype of which is presented in Figure~\ref{fig:m3uphoto}, while Figure~\ref{fig:m3udesign} provides an overview of its components. Recently published work employs a perception unit that relies on Near Infrared cameras and inertial sensors for localization~\cite{NIRdepth_ICUAS_2017}.

%
%

%
%

\begin{figure}[h!]
\centering
  \includegraphics[width=0.99\columnwidth]{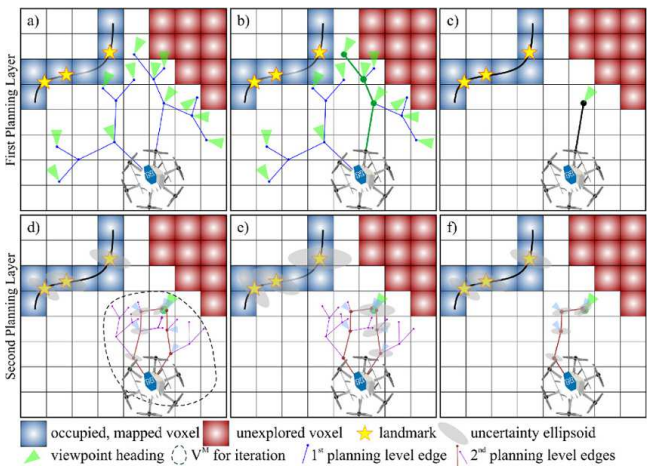}
\caption{2D representation of the two--steps uncertainty--aware exploration and mapping planner. The first planning layer samples the path with the maximum exploration gain. The viewpoint configuration of the first vertex of this path becomes the reference to the second planning layer. Then this step, samples admissible paths that arrive to this configuration, performs belief propagation along the tree edges, and selects the one that provides minimum uncertainty over the robot pose and tracked landmarks.  }
\label{fig:rhem_steps}
\end{figure}

With the localization pipeline running onboard the robot, the robot pose and tracked landmarks as well as their covariance matrix are estimated. These estimates are then exploited from the path planning module and propagated along sampled paths in order to account for the robot localizability along different trajectories. Figure~\ref{fig:rhem_steps} presents the localizability--aware exploration and mapping planner~\cite{RHEM_ICRA_2017}. At first, in an online computed tree, the algorithm identifies the branch that optimizes the amount of new space expected to be explored. The first viewpoint configuration of this branch is selected, but the path towards it is decided through a second planning step. Within that, a new random tree is sampled, admissible branches arriving at the reference viewpoint are found and the robot belief about its state and the tracked landmakrs of the environment is propagated. As system state the concatenation of the robot states and tracked landmarks (features) is considered. Then, the branch that minimizes the localization uncertainty, as factorized using the D--optimality of the pose and landmarks covariance matrix is selected. The corresponding path is conducted by the robot and the process is iteratively repeated. It is noted that this process goes beyond baseline deterministic exploration~\cite{NBVP_ICRA_16,yoder2016autonomous}. When some knowledge of the environment is available as a prior map, work on optimized coverage can also be exploited to provide a rough global path~\cite{SIP_AURO_2015,APST_MSC_2015,bircher_robotica,BABOOMS_ICRA_15,galceran2013survey,papachristos2016distributed,oettershagen2016long,alexis2017realizing}.

%
%

\section{ROBOTIC RADIATION DETECTION}\label{sec:rad}

\begin{figure}[h!]
\centering
  \includegraphics[width=0.99\columnwidth]{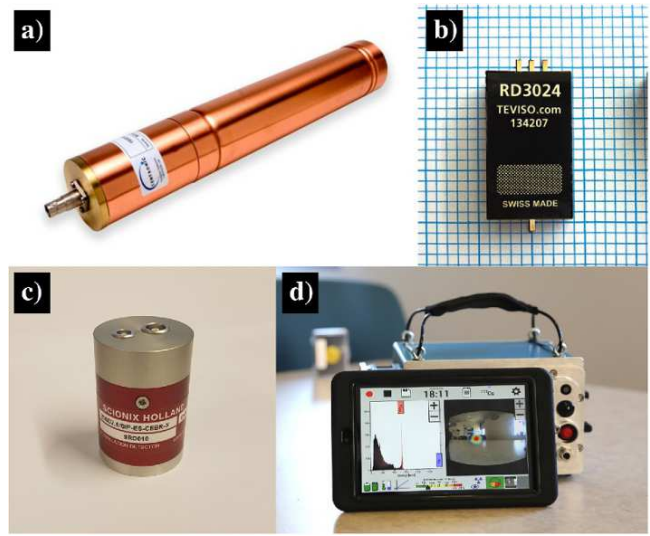}
\caption{Indicative radiation detectors: a) gas--filled proportional detector, b) solid--state detector, c) scintillator, d) gamma camera.}
\label{fig:raddetectors}
\end{figure}

Radiation detection is a well--studied and a continuously--evolving field on its own but robotized sensing brings further and new challenges. First of all, good overview of the types of radiation sensing systems, such as proportional gas--filled detectors, semiconductor diode detectors, germanium gamma-ray detectors and other solid--state solutions, scintillation detectors, and radiation cameras, their features, radiation, thermal, and mechanical hardness is required. Furthermore, the critical role of photomultiplier tubes and photodiodes has to be well--understood to enable the appropriate selection and design of the sensing module. Good theoretical references can be found at~\cite{knoll2010radiation,tsoulfanidis2013measurement}. Specific to the application, the sensing solution has to be decided according to the interest to detect alpha, beta, gamma or neutron activity, the required energy resolution and the power levels of the site to be surveyed. A critical question is if spectroscopy is required. Figure~\ref{fig:raddetectors} presents indicative radiation detectors. In addition, limitations of the robotic platform will necessarily shape the final detector selection.

%
%

In the area of gamma radiation detection, and depending on the application, three detection technologies namely a) miniature scintillation detectors (e.g. CeBr3, CsI, NaI) with built--in temperature compensated bias generator and a pre-amplifier often alongside a silicon photomultiplier (SiPm) tube~\cite{quarati2013scintillation,swiderski2015gamma,valtonen2009radiation,beck1972situ,buzhan2003silicon,herbert2006first}, b) miniature solid--state low voltage gamma detectors~\cite{dearnaley1967solid,lutz1999semiconductor}, and c) gamma cameras~\cite{he2010three,bolotnikov2012array,zhang2007prototype} are worth of special attention. The first two solutions can be realized at extremely small sizes and low--weights making them affordable for aerial robotic applications, while scintillation devices can provide the sensitivity and energy resolution characteristics required for precise monitoring and source localization. Radiation cameras are still relatively heavy but provide unique characteristics when it comes to radiation mapping in correlation with the $3\textrm{D}$ structure. Through a multi--modal sensor fusion approach, comprehensive 3D maps annotated with radiation can be derived. 

Neutron detection is also a particularly interesting area with high relevance to homeland security and industrial monitoring (e.g. personnel monitoring, water content in soil) applications. Neutron detection refers to the effective detection of neurons entering a well--positioned detector. Neutrons can be produced through multiple processes such as alpha particle induced reactions, spontaneous fission, and induced fission. Gas--filled proportional detectors such as the family of $^3$He--based detectors~\cite{east1969polyethylene}, scintillation neutron detectors (e.g. liquid organic, plastic)~\cite{thomas1962boron}, as well as solid--state neutron detectors may be used~\cite{petrillo1996solid}. A selection of a neutron detector with the appropriate radiological sensitivity for the application is required.
  
Alpha detection is key to many applications in contaminated areas but its detection is particularly challenging. As alpha particles are the heaviest and most highly charged of the common nuclear radiations, they quickly give up their energy to any medium through which they pass, rapidly coming to equilibrium with, and disappearing in the medium. Due to this reason special detection techniques must be used to allow the particles to enter the active region of a detector (e.g. ZnS(Ag)--based scintillation devices)). In field instruments it is common to use an extremely thin piece of aluminized Mylar film on the face of the detector probe to cover a thin layer of florescent material. This is due to the fact that energy attenuation of the incident alpha radiation through Mylar is estimated to be less than $10\%$. However, the use of this film makes the detector extremely fragile - to the level that any contact with a hard object, such as a blade of hard grass, may puncture the film~\cite{websiterad}. 

Beyond the radiation detectors themselves, a set of methods and techniques are critical to achieve the desired final sensing result. First of all, spectroscopy is critical when characterization matters. Dose and dose--rate equivalent count rate monitoring is important especially for safety--related tasks. Facilitation and tuning of detection directionality through a set of techniques (e.g. compton imaging, coded mask apertures, collimation) allows to realize the desired sensing properties. Furthermore, appropriate design of the interfacing (e.g. amplifiers, multi--channel analyzers, analog--to--digital converters) and processing electronics (e.g. DSPs, FPGAs) is critical and has to be considered in order to achieve the desired sensing functionality at an affordable weight and cost. 

\begin{figure}[h!]
\centering
  \includegraphics[width=0.99\columnwidth]{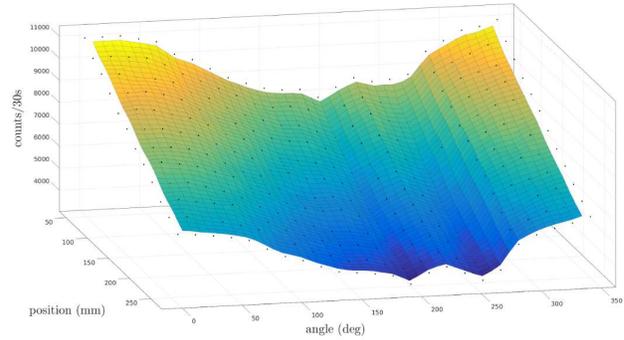}
\caption{Calibration results of a TEVISO RD3024 solid--state detector with the use of a $300\textrm{mCi}$ $\textrm{Cs}$-127 source. Calibration took place over different sensor orientations and varied distances from the source. }
\label{fig:teviso_calib}
\end{figure}

An additional critical step is that of detector calibration. The process of radiation detector intrinsics calibration involves the use of a pre--calibrated source and logging of counting statistics over different orientations and distances from the source. It is important to be aware of the polarity characteristics of the radiation detector to be used and ongoing experience indicates that an in--house calibration step is critical to take place. Furthermore, extrinsics calibration with the remaining of the sensor modalities is again required if correlation with the $3\textrm{D}$ reconstructed map or direct sensor data (e.g. camera frames) is to take place. Figure~\ref{fig:teviso_calib} presents such calibration data for the case of the TEVISO RD3024 solid--state detector with the use of a $300\textrm{mCi}$ $\textrm{Cs}$-127 source.  

%
%

Finally, a challenge of robotized nuclear detection especially related to platforms of high mobility (e.g. aerial robots) is that of localization accuracy. Due to the fact that depending on the application radiation detectors may require significant dwell times, localization accuracy and robustness is critical. This is particularly relevant when smaller spaces are considered, when GPS--denied operation is required, and when accurate estimates of the radiation source location are required. 

\section{PRELIMINARY RESULTS}\label{sec:results}

A set of preliminary studies related to exploration in DVEs and radiation detection have been conducted in order to approach the challenge of developing nuclearized robotics especially related with the problem of supporting the waste management decommissioning effort. 

\subsection{Localization and Mapping inside a Dark Tunnel}
For this experimental evaluation, the mission took place within a remote city tunnel during night--time. This kind of environment is unique in multiple aspects: a) it exhibits nearly--complete lack of ambient light due to its closed structure especially at night (while even during the day it still is significantly dark and robotic deployment within such a space would still require handling of this aspect), b) it is littered with dust which can lift up into the air and into the sensors' fields--of--view due to the turbulence created by an aerial robot's spinning rotors, and c) its internal structure mainly composed of concrete walls is such that contains little discernible texture. 

%
\begin{figure}[h!]
\centering
  \includegraphics[width=0.99\columnwidth]{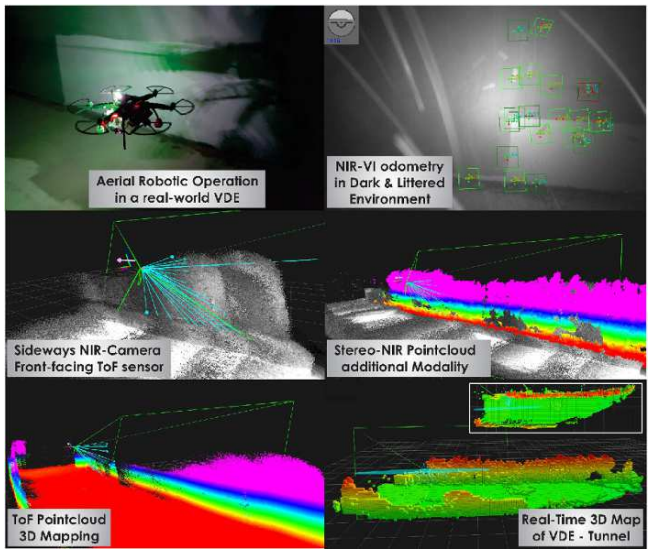}
\caption{Aerial robotic localization and mapping within a dark city tunnel.}
\label{fig:tunnel_experiment}
\end{figure}
%

Figure~\ref{fig:tunnel_experiment} illustrates these conditions based on the data recorded during the experiment, alongside the localization and mapping results as performed in real--time during the experiment. A video of the experimental sequence is also available at \url{https://youtu.be/HpWlFUNboR4}

\subsection{Autonomous Exploration in DVEs}
This mission scenario refers to the complete concept of autonomous robotic exploration of DVEs. The mock-up space is a dark indoor location, with dimensions $12\times6.5\times2\textrm{m}$, setup to incorporate artificially--created vertical and T--shaped walls, as well as other structural elements by using $300$ boxes with size $0.4\times0.3\times0.3\textrm{m}$. 

%
\begin{figure}[h!]
\centering
  \includegraphics[width=0.99\columnwidth]{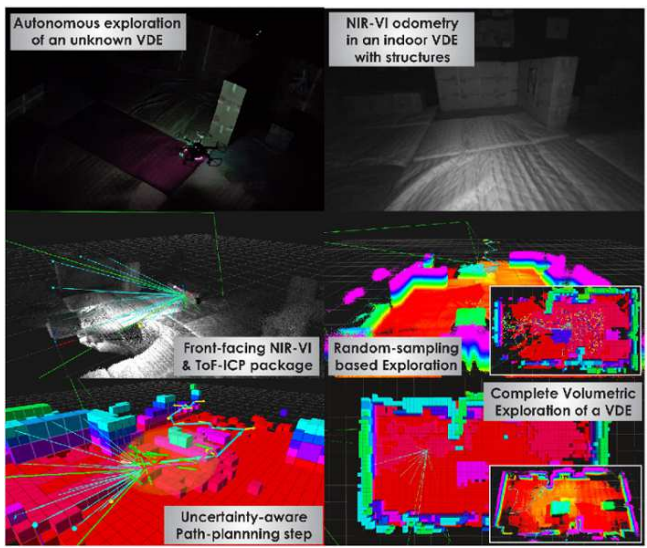}
\caption{Autonomous aerial robotic exploration within an indoor degraded visual environment containing wall-like structures.}
\label{fig:arena_experiment}
\end{figure}
%

Figure~\ref{fig:arena_experiment} illustrates the aforementioned conditions, as well as the progress of this experiment, while a video of the sequence is also available at \url{https://youtu.be/1-nPFBhyTBM}. For this mission where the human is out--of--the--loop, consistent localization and mapping during autonomous exploration are provided by the localizability-aware RHEM planner. 

\subsection{Radiation Detection}
\begin{figure}[h!]
\centering
  \includegraphics[width=0.99\columnwidth]{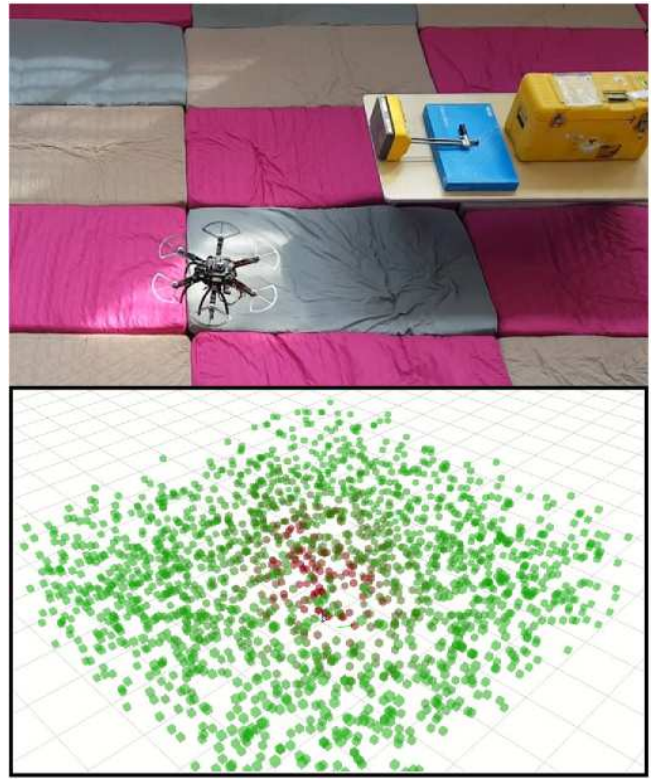}
\caption{Aerial robotic radiation detection.}
\label{fig:radexp}
\end{figure}

Two Teviso RD3024 low voltage SMD/SMT nuclear radiation sensors were installed on the aerial robot. As mounting points the two opposite facing arms of its hexacopter structure were selected to create differential measurements and exploit the polarity of the sensor. With the radiation sensors being initially calibrated with the use of a characterized source (see Figure~\ref{fig:teviso_calib}), the robot was then commanded to follow an exploration trajectory. Given the sensor data collected, the source location is estimated. Figure~\ref{fig:radexp} presents the relevant result, while a video of the experiment is available at \url{https://youtu.be/b9BbKQTfrY8}.

%
%

\section{CONCLUSIONS}\label{sec:concl}

This paper discussed certain challenges relevant to the problem of robotically supported nuclear waste management and decommissioning with a special focus on nuclear site characterization. Furthermore, preliminary results on GPS--denied operation in degraded visual environments, exploration and mapping, as well as radiation detection were presented. Future work will focus on the challenges of multi--modal characterization, robustly autonomous exploration and mapping, optimized robotic radiation detection, and real--time multi--source localization.

\section{ACKNOWLEDGEMENTS}\label{sec:concl}

The authors would like to thank Myung Chul Jo, radiation safety officer at the University of Nevada, Reno who provided his valuable help in order to enable the proper and safe use of the radiation equipment involved in this study.


\begin{thebibliography}{10}
\providecommand{\url}[1]{#1}
\csname url@samestyle\endcsname
\providecommand{\newblock}{\relax}
\providecommand{\bibinfo}[2]{#2}
\providecommand{\BIBentrySTDinterwordspacing}{\spaceskip=0pt\relax}
\providecommand{\BIBentryALTinterwordstretchfactor}{4}
\providecommand{\BIBentryALTinterwordspacing}{\spaceskip=\fontdimen2\font plus
\BIBentryALTinterwordstretchfactor\fontdimen3\font minus
  \fontdimen4\font\relax}
\providecommand{\BIBforeignlanguage}[2]{{%
\expandafter\ifx\csname l@#1\endcsname\relax
\typeout{** WARNING: IEEEtran.bst: No hyphenation pattern has been}%
\typeout{** loaded for the language `#1'. Using the pattern for}%
\typeout{** the default language instead.}%
\else
\language=\csname l@#1\endcsname
\fi
#2}}
\providecommand{\BIBdecl}{\relax}
\BIBdecl

\bibitem{qian2012small}
K.~Qian, A.~Song, J.~Bao, and H.~Zhang, ``Small teleoperated robot for nuclear
  radiation and chemical leak detection,'' \emph{International Journal of
  Advanced Robotic Systems}, vol.~9, no.~3, p.~70, 2012.

\bibitem{han2013low}
J.~Han, Y.~Xu, L.~Di, and Y.~Chen, ``Low-cost multi-uav technologies for
  contour mapping of nuclear radiation field,'' \emph{Journal of Intelligent \&
  Robotic Systems}, pp. 1--10, 2013.

\bibitem{bogue2011robots}
R.~Bogue, ``Robots in the nuclear industry: a review of technologies and
  applications,'' \emph{Industrial Robot: An International Journal}, vol.~38,
  no.~2, pp. 113--118, 2011.

\bibitem{cortez2009distributed}
R.~A. Cortez, H.~G. Tanner, and R.~Lumia, ``Distributed robotic radiation
  mapping,'' in \emph{Experimental Robotics}.\hskip 1em plus 0.5em minus
  0.4em\relax Springer, 2009, pp. 147--156.

\bibitem{dudar1999mobile}
A.~M. Dudar, C.~R. Ward, J.~D. Jones, W.~R. Mallet, L.~J. Harpring, M.~X.
  Collins, and E.~K. Anderson, ``Mobile autonomous robotic apparatus for
  radiologic characterization,'' 1999, uS Patent 5,936,240.

\bibitem{westrom1994radiation}
G.~B. Westrom, R.~E. Carlton, and L.~R. Tripp, ``Radiation mapping system,''
  1994, uS Patent 5,286,973.

\bibitem{michal2009red}
R.~Michal, ``Red whittaker and the robots that helped clean uptmi-2,''
  \emph{Nuclear News}, pp. 37--40, 2009.

\bibitem{groves1983now}
L.~R. Groves, \emph{Now it can be told: The story of the Manhattan
  Project}.\hskip 1em plus 0.5em minus 0.4em\relax Da Capo Press, 1983.

\bibitem{kelly2009manhattan}
C.~C. Kelly, \emph{Manhattan Project: The Birth of the Atomic Bomb in the Words
  of Its Creators, Eyewitnesses, and Historians}.\hskip 1em plus 0.5em minus
  0.4em\relax Black Dog \& Leventhal, 2009.

\bibitem{gephart2003hanford}
R.~E. Gephart, ``Hanford: A conversation about nuclear waste and cleanup,''
  Pacific Northwest National Laboratory (PNNL), Richland, WA (US), Tech. Rep.,
  2003.

\bibitem{tsubokura2012internal}
M.~Tsubokura, S.~Gilmour, K.~Takahashi, T.~Oikawa, and Y.~Kanazawa, ``Internal
  radiation exposure after the fukushima nuclear power plant disaster,''
  \emph{Jama}, vol. 308, no.~7, pp. 669--670, 2012.

\bibitem{hopkinson2004radiation}
G.~R. Hopkinson, A.~Mohammadzadeh, and R.~Harboe-Sorensen, ``Radiation effects
  on a radiation-tolerant cmos active pixel sensor,'' \emph{IEEE Transactions
  on Nuclear Science}, vol.~51, no.~5, pp. 2753--2762, 2004.

\bibitem{goiffon2008ionizing}
V.~Goiffon, P.~Magnan, F.~Bernard, G.~Rolland, O.~Saint-P{\'e}, N.~Huger, and
  F.~Corbi{\`e}re, ``Ionizing radiation effects on cmos imagers manufactured in
  deep submicron process,'' in \emph{Proceedings of SPIE-The International
  Society for Optical Engineering}, vol. 6816, 2008.

\bibitem{jarman2011bayesian}
K.~D. Jarman, E.~A. Miller, R.~S. Wittman, and C.~J. Gesh, ``Bayesian radiation
  source localization,'' \emph{Nuclear technology}, vol. 175, no.~1, pp.
  326--334, 2011.

\bibitem{towler2012radiation}
J.~Towler, B.~Krawiec, and K.~Kochersberger, ``Radiation mapping in
  post-disaster environments using an autonomous helicopter,'' \emph{Remote
  Sensing}, vol.~4, no.~7, pp. 1995--2015, 2012.

\bibitem{morelande2009radiological}
M.~R. Morelande and B.~Ristic, ``Radiological source detection and localisation
  using bayesian techniques,'' \emph{IEEE Transactions on Signal Processing},
  vol.~57, no.~11, pp. 4220--4231, 2009.

\bibitem{christie2016radiation}
G.~Christie, A.~Shoemaker, K.~Kochersberger, P.~Tokekar, L.~McLean, and
  A.~Leonessa, ``Radiation search operations using scene understanding with
  autonomous uav and ugv,'' \emph{arXiv preprint arXiv:1609.00017}, 2016.

\bibitem{NIRdepth_ICUAS_2017}
{C. Papachristos, S. Khattak, and K. Alexis}, ``Autonomous exploration of
  visually-degraded environments using aerial robots,'' in \emph{IEEE
  International Conference on Unmanned Aircraft Systems (ICUAS)}, 2017.

\bibitem{RHEM_ICRA_2017}
S.~K. Christos~Papachristos and K.~Alexis, ``Uncertainty--aware receding
  horizon exploration and mapping using aerial robots,'' in \emph{IEEE
  International Conference on Robotics and Automation (ICRA)}, June 2017.

\bibitem{NBVP_ICRA_16}
{A. Bircher, M. Kamel, K. Alexis, H. Oleynikova and R. Siegwart}, ``Receding
  horizon "next-best-view" planner for 3d exploration,'' in \emph{IEEE
  International Conference on Robotics and Automation (ICRA)}, May 2016.

\bibitem{yoder2016autonomous}
L.~Yoder and S.~Scherer, ``Autonomous exploration for infrastructure modeling
  with a micro aerial vehicle,'' in \emph{Field and Service Robotics}.\hskip
  1em plus 0.5em minus 0.4em\relax Springer, 2016, pp. 427--440.

\bibitem{SIP_AURO_2015}
{A. Bircher, M. Kamel, K. Alexis, M. Burri, P. Oettershagen, S. Omari, T.
  Mantel and R. Siegwart}, ``\BIBforeignlanguage{English}{Three-dimensional
  coverage path planning via viewpoint resampling and tour optimization for
  aerial robots},'' \emph{\BIBforeignlanguage{English}{Autonomous Robots}}, pp.
  1--25, 2015.

\bibitem{APST_MSC_2015}
{K. Alexis, C. Papachristos, R. Siegwart, and A. Tzes}, ``Uniform coverage
  structural inspection path-planning for micro aerial vehicles,'' September
  2015.

\bibitem{bircher_robotica}
{A. Bircher, K. Alexis, U. Schwesinger, S. Omari, M. Burri, and R. Siegwart},
  ``An incremental sampling-based approach to inspection planning: The
  rapidly-exploring random tree of trees,'' 2015.

\bibitem{BABOOMS_ICRA_15}
{A. Bircher, K. Alexis, M. Burri, P. Oettershagen, S. Omari, T. Mantel and R.
  Siegwart}, ``Structural inspection path planning via iterative viewpoint
  resampling with application to aerial robotics,'' in \emph{IEEE International
  Conference on Robotics and Automation (ICRA)}, May 2015, pp. 6423--6430.

\bibitem{galceran2013survey}
E.~Galceran and M.~Carreras, ``A survey on coverage path planning for
  robotics,'' \emph{Robotics and Autonomous Systems}, vol.~61, no.~12, pp.
  1258--1276, 2013.

\bibitem{papachristos2016distributed}
C.~Papachristos, K.~Alexis, L.~R.~G. Carrillo, and A.~Tzes, ``Distributed
  infrastructure inspection path planning for aerial robotics subject to time
  constraints,'' in \emph{Unmanned Aircraft Systems (ICUAS), 2016 International
  Conference on}.\hskip 1em plus 0.5em minus 0.4em\relax IEEE, 2016, pp.
  406--412.

\bibitem{oettershagen2016long}
P.~Oettershagen, T.~Stastny, T.~Mantel, A.~Melzer, K.~Rudin, P.~Gohl,
  G.~Agamennoni, K.~Alexis, and R.~Siegwart, ``Long-endurance sensing and
  mapping using a hand-launchable solar-powered uav,'' in \emph{Field and
  Service Robotics}.\hskip 1em plus 0.5em minus 0.4em\relax Springer
  International Publishing, 2016, pp. 441--454.

\bibitem{alexis2017realizing}
K.~Alexis, ``Realizing the aerial robotic worker for inspection operations,''
  \emph{arXiv preprint arXiv:1703.02640}, 2017.

\bibitem{knoll2010radiation}
G.~F. Knoll, \emph{Radiation detection and measurement}.\hskip 1em plus 0.5em
  minus 0.4em\relax John Wiley \& Sons, 2010.

\bibitem{tsoulfanidis2013measurement}
N.~Tsoulfanidis, \emph{Measurement and detection of radiation}.\hskip 1em plus
  0.5em minus 0.4em\relax CRC press, 2013.

\bibitem{quarati2013scintillation}
F.~Quarati, P.~Dorenbos, J.~Van~der Biezen, A.~Owens, M.~Selle, L.~Parthier,
  and P.~Schotanus, ``Scintillation and detection characteristics of
  high-sensitivity cebr 3 gamma-ray spectrometers,'' \emph{Nuclear Instruments
  and Methods in Physics Research Section A: Accelerators, Spectrometers,
  Detectors and Associated Equipment}, vol. 729, pp. 596--604, 2013.

\bibitem{swiderski2015gamma}
L.~Swiderski, P.~Schotanus, E.~Bodewits, D.~Badocco, T.~Batsch, D.~Cester,
  M.~Corbo, P.~Garosi, A.~Iovene, J.~Iwanowska-Hanke \emph{et~al.}, ``Gamma
  spectrometer based on cebr 3 scintillator with compton suppression for
  identification of trace activities in water,'' in \emph{Nuclear Science
  Symposium and Medical Imaging Conference (NSS/MIC), 2015 IEEE}.\hskip 1em
  plus 0.5em minus 0.4em\relax IEEE, 2015, pp. 1--3.

\bibitem{valtonen2009radiation}
E.~Valtonen, J.~Peltonen, O.~Dudnik, A.~Kudin, H.~Andersson, Y.~A. Borodenko,
  T.~Eronen, J.~Huovelin, H.~Kettunen, E.~Kurbatov \emph{et~al.}, ``Radiation
  tolerance tests of small-sized csi (tl) scintillators coupled to
  photodiodes,'' \emph{IEEE Transactions on Nuclear Science}, vol.~56, no.~4,
  pp. 2149--2154, 2009.

\bibitem{beck1972situ}
H.~L. Beck, J.~Decampo, and C.~Gogolak, ``In situ ge (li) and nai (tl)
  gamma-ray spectrometry.'' Health and Safety Lab.,(AEC), New York, Tech. Rep.,
  1972.

\bibitem{buzhan2003silicon}
P.~Buzhan, B.~Dolgoshein, L.~Filatov, A.~Ilyin, V.~Kantzerov, V.~Kaplin,
  A.~Karakash, F.~Kayumov, S.~Klemin, E.~Popova \emph{et~al.}, ``Silicon
  photomultiplier and its possible applications,'' \emph{Nuclear Instruments
  and Methods in Physics Research Section A: Accelerators, Spectrometers,
  Detectors and Associated Equipment}, vol. 504, no.~1, pp. 48--52, 2003.

\bibitem{herbert2006first}
D.~J. Herbert, V.~Saveliev, N.~Belcari, N.~D'Ascenzo, A.~Del~Guerra, and
  A.~Golovin, ``First results of scintillator readout with silicon
  photomultiplier,'' \emph{IEEE Transactions on Nuclear Science}, vol.~53,
  no.~1, pp. 389--394, 2006.

\bibitem{dearnaley1967solid}
G.~Dearnaley, ``Solid-state radiation detectors,'' \emph{Contemporary Physics},
  vol.~8, no.~6, pp. 607--626, 1967.

\bibitem{lutz1999semiconductor}
G.~Lutz \emph{et~al.}, \emph{Semiconductor radiation detectors}.\hskip 1em plus
  0.5em minus 0.4em\relax Springer, 1999, vol.~40.

\bibitem{he2010three}
Z.~He and F.~Zhang, ``Three-dimensional, position-sensitive radiation
  detection,'' 2010, uS Patent 7,692,155.

\bibitem{bolotnikov2012array}
A.~Bolotnikov, J.~Butcher, G.~Camarda, Y.~Cui, G.~De~Geronimo, J.~Fried,
  R.~Gul, P.~Fochuk, M.~Hamade, A.~Hossain \emph{et~al.}, ``Array of virtual
  frisch-grid czt detectors with common cathode readout for correcting charge
  signals and rejection of incomplete charge-collection events,'' \emph{IEEE
  Transactions on Nuclear Science}, vol.~59, no.~4, pp. 1544--1551, 2012.

\bibitem{zhang2007prototype}
F.~Zhang, Z.~He, and C.~E. Seifert, ``A prototype three-dimensional position
  sensitive cdznte detector array,'' \emph{IEEE Transactions on Nuclear
  Science}, vol.~54, no.~4, pp. 843--848, 2007.

\bibitem{east1969polyethylene}
L.~East and R.~Walton, ``Polyethylene moderated 3he neutron detectors,''
  \emph{Nuclear Instruments and Methods}, vol.~72, no.~2, pp. 161--166, 1969.

\bibitem{thomas1962boron}
G.~Thomas, ``A boron-loaded liquid scintillation neutron detector using a
  single photomultiplier,'' \emph{Nuclear Instruments and Methods}, vol.~17,
  no.~2, pp. 137--139, 1962.

\bibitem{petrillo1996solid}
C.~Petrillo, F.~Sacchetti, O.~Toker, and N.~Rhodes, ``Solid state neutron
  detectors,'' \emph{Nuclear Instruments and Methods in Physics Research
  Section A: Accelerators, Spectrometers, Detectors and Associated Equipment},
  vol. 378, no.~3, pp. 541--551, 1996.

\bibitem{websiterad}
{Office of the Assistant Secretary of Defense for Nuclear, Chemical, and
  Biological Defense Programs/Nuclear Materials}, ``Radiation detection and
  measurement.''

\end{thebibliography}
\end{document}